\documentclass[a4paper,fleqn]{cas-sc}

\usepackage[numbers]{natbib}
\usepackage{graphicx}
\usepackage{subfigure}
\usepackage{amsmath}
\usepackage{amsfonts}
\usepackage{setspace}
\usepackage{multirow}
\usepackage{booktabs}

\def\tsc#1{\csdef{#1}{\textsc{\lowercase{#1}}\xspace}}
\tsc{WGM}
\tsc{QE}
\tsc{EP}
\tsc{PMS}
\tsc{BEC}
\tsc{DE}
\begin{document}
\let\WriteBookmarks\relax
\def\floatpagepagefraction{1}
\def\textpagefraction{.001}
\shorttitle{$\omega$-Net: Dual Supervised Medical Image Segmentation Model with MDA and Cascade MSC}
\shortauthors{Bo Wang et~al.}
%\begin{frontmatter}

\title [mode = title]{$\omega$-Net: Dual Supervised Medical Image Segmentation Model with Multi-Dimensional Attention and Cascade Multi-Scale Convolution}

\author[1,2]{Bo Wang}

\author[1,2]{Lei Wang}
\cormark[1]
\author[3]{Junyang Chen}

\author[1,2]{Zhenghua Xu}
\cormark[2]
\ead{zhenghua.xu@hebut.edu.cn}

\author[4]{Thomas Lukasiewicz}

\author[5]{Zhigang Fu}

\address[1]{State Key Laboratory of Reliability and Intelligence of Electrical Equipment, Hebei University of Technology, China.}

\address[2]{Tianjin Key Laboratory of Bioelectromagnetic Technology and Intelligent Health, Hebei University of Technology, China.}

\address[3]{Department of Computer Science, University of Macau, China.}

\address[4]{Department of Computer Science, University of Oxford, United Kingdom.}

\address[5]{Department of Health Management Center, 983 Hospital of Joint Logistics Support Force, China.}

\cortext[cor1]{Co-first author}
\cortext[cor2]{Corresponding author}

\begin{abstract}
%$\omega$-net.
Deep learning based medical image segmentation technology aims at automatic recognizing and annotating objects on the medical image. Non-local attention and feature learning by multi-scale methods are widely used to model network, which drives progress in medical image segmentation. However, those attention mechanism methods have weakly non-local receptive fields' strengthened connection for small objects in medical images. Then, the features of important small objects in abstract or coarse feature maps may be deserted, which leads to unsatisfactory performance. Moreover, the existing multi-scale methods only simply focus on different sizes of view, whose sparse multi-scale features collected are not abundant enough for small objects segmentation. In this work, a multi-dimensional attention segmentation model with cascade multi-scale convolution is proposed to predict accurate segmentation for small objects in medical images. As the weight function, multi-dimensional attention modules provide coefficient modification/strengthening for significant/informative small objects features. Furthermore, The cascade multi-scale convolution modules in each skip-connection path are exploited to capture multi-scale features in different semantic depth. The proposed method is evaluated on three datasets: KiTS19, Pancreas CT of Decathlon-10, and MICCAI 2018 LiTS Challenge, demonstrating better segmentation performances than the state-of-the-art baselines.
\end{abstract}


\maketitle

\section{Introduction}
With the fast development of deep learning, deep learning based medical image analysis technologies have been increasingly applied in clinical computer-aided diagnosis (CAD)~\cite{gibson2018niftynet,hu2018deep}. Deep learning based medical image segmentation is one of the important tasks~\cite{dalca2018anatomical} in CAD, which aims to recognize and annotate the interested regions with masks and/or outlines using deep models. U-Net~\cite{ronneberger2015u} is a widely exploited deep learning based medical image segmentation model. In addition, to further improve the segmentation performances, many variants of U-Net have been proposed in recent works~\cite{zhou2018unet++}. Attention U-Net~\cite{oktay2018attention} integrates attention gates into the expansive path to suppress the response of the irrelevant background information and enhance the sensitivity of foreground information.

Although U-Net and its variants have already achieved some great successes, their segmentation accuracies for small objects are still unsatisfactory. Although skip connections are used to remedy this problem, using solely skip connections is still not sufficient to fully recover the lost information, and will inevitably results in inaccurate segmentation for small objects. Therefore, the need of a more accurate deep model for medical image segmentation is compelling. Up-sampled feature maps are obtained by applying an up-sampling operation on feature maps in the previous layer; while attentive feature maps are generated by a \emph{multi-dimensional attention module}.

We further introduce additional \emph{cascade multi-scale convolution} (Cascade MSC) operations on the skip-connection and MDA module of MDA-U-Net. Experiments show that Cascade MSC operations can further enhance the model's segmentation performance on medical images.

The contributions of this paper are briefly summarized as follows:

\begin{itemize}
	\item We point out the limitation of the existing U-Net based deep models in small object segmentation, and propose a novel $\omega$-Net model to resolve this problem by using multi-dimensional attention module to strengthen the deep model's learning capability on areas and channels with important small object features.
	\item In addition, cascade multi-scale convolution (Cascade MSC) operations are introduced to further enhance the deep model's segmentation performances in medical image.
\end{itemize}

\section{Related Work}
\label{Related Work}
Deep learning-based methods have already been successfully applied in medical image segmentation. FCN~\cite{long2015fully} sums the feature maps of up-sampling with feature maps skipped from the encoder to produce semantic segmentation by pixel-level classification. Based on U-Net, U-Net$++$~\cite{zhou2018unet++} exploit a series of nested dense skip pathways to reduce the semantic gap between the feature maps of the contracting path and expanding path.

\textbf{Attention Mechanisms.} In recent years, some other works have also utilized attention mechanisms to improve the performance of deep learning-based segmentation models. Therefore, Wang et al.~\cite{wang2018non} compute the weighted sum of the features at all positions to capture long-range dependencies for the global receptive field. Nevertheless, the non-local block spends a high calculated cost to take the weights of all the position features into consideration. Furthermore, Dual Attention Networks (DANet)~\cite{fu2019dual} is proposed to integrate local features with their global dependencies adaptively. They append two types of attention modules on top of traditional dilated FCN, which model the semantic inter-dependencies in spatial and channel dimensions, respectively.

\textbf{Multi-Scale Methods.} Furthermore, as for multi-scale pooling or convolutional operations, the architectural decisions were designed based on the Hebbian principle and the intuition of multi-scale processing~\cite{szegedy2015going} to optimize the quality of features extraction. Moreover, Zhao et al.~\cite{zhao2017pyramid} exploited the global context information through the pyramid pooling module and the proposed pyramid scene parsing network (PSPNet).

Overall, most of the existing studies focus on the encoding-decoding models that use attention mechanisms or multi-scale methods to predict segmentation from end-to-end. Our work is built based on U-Net, and the novelty lies in the learning of dense multi-scale weighted features by multi-dimensional attention module and cascade multi-scale convolution module. Our model is flexible to incorporate rich context constraint information.

%%%%%%%%%%%%%%%%%%%%%%%%%%%%%%%%%%%%
%\begin{figure*}
%\vspace{-1mm}
%\centering
%\includegraphics[width=\linewidth]{figs/fig_01}
%\caption{The construction of the U-Net with multi-dimensional attention (MDA-U-Net). MDA-U-Net integrates the contracting path, an expansive path with auxiliary loss, and an additional expansive path equipped with the multi-dimensional attention module.}
%\label{fig 1}
%\vspace{-1mm}
%\end{figure*}
%%%%%%%%%%%%%%%%%%%%%%%%%%%%%%%%%%%%
\begin{figure*}
	\vspace{-1mm}
	\centering
	\includegraphics[width=\linewidth]{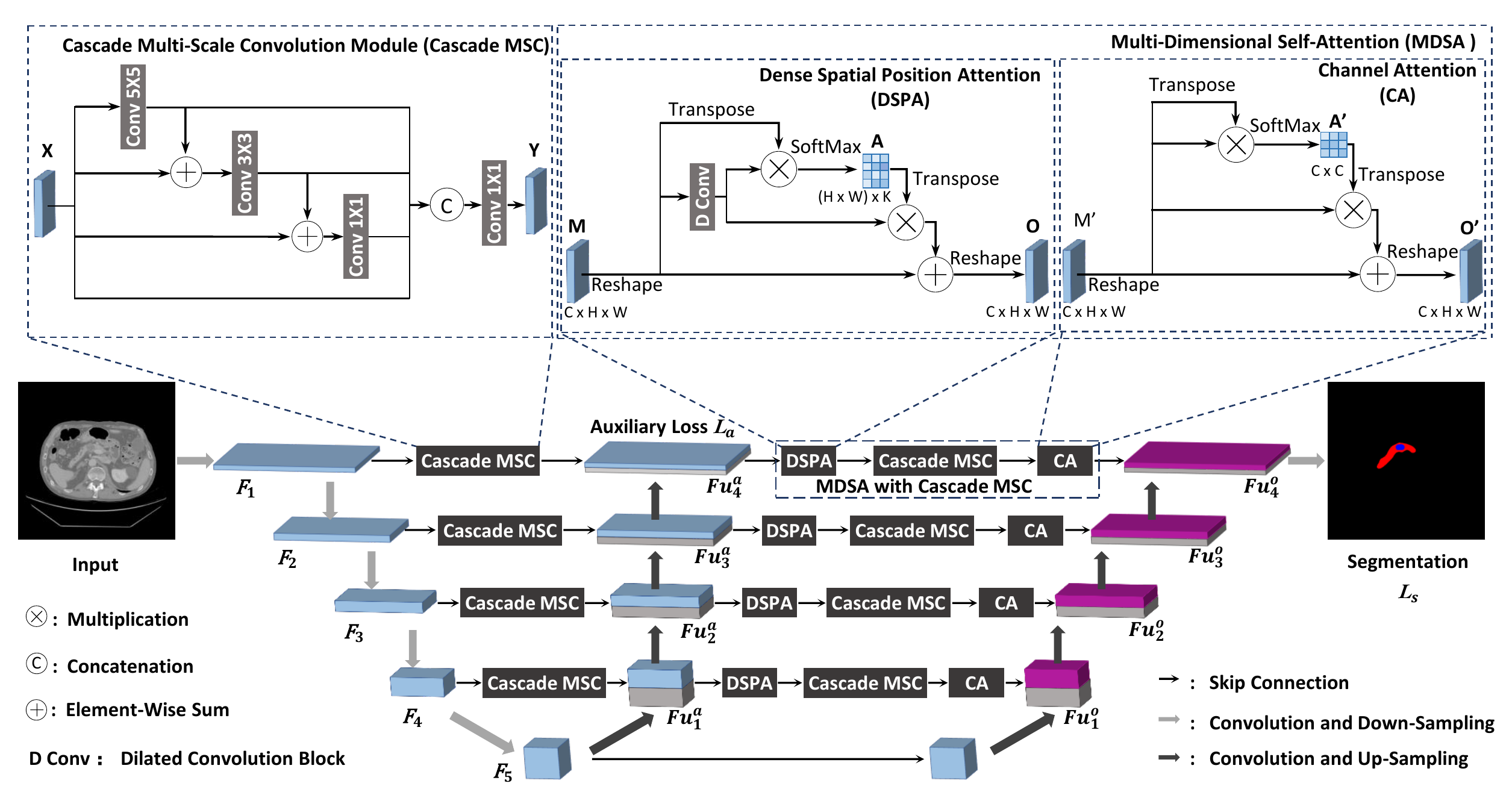}
	\caption{Overall structure of \emph{$\omega$-Net}}
	\label{fig_1}
	\vspace{-1mm}
\end{figure*}

%\section{Methods}
%\label{Methods}
%In this section, we introduce the algorithm formula, and present in detail our proposed method, followed by the descriptions of \emph{MDA-U-Net} firstly shown in Fig.~\ref{fig 1} (Section \ref{sec_DMA-U-Net}), \emph{MDA$^*$-U-Net$^*$} further illustrated in Fig.~\ref{fig 2} (Section \ref{sec_DMA-U-Net$++$}), and the loss function for training the networks (Section \ref{sec_Loss Function}).  The architecture of our proposed model is detailed as follows.

\section{Dual Supervised Medical Image Segmentation Model with MDA and Cascade MSC}
\label{sec_DMA-U-Net}
%2nd paragraph main idea: Using solely skip connections may not be sufficent to fully recover the lost information, and inevitably results in inaccurate segmentation for small objects. These minor errors may be tolerable in natural image segmentation, but are unacceptable in medical image segmentation, because they may cuase fatal consequences in clinical practice (e.g., when this model is used to delineate the target area of ??tumor radiotherapy, even a few tumor cells missed may cause failure of radiotherapy and cancer recurrence). Therefore, the need of a more accurate deep model for medical image segmentation is compelling.

%3nd paragraph main idea: Therefore, in this paper, we propose a novel $\omega$-net model for more accurate medical image segmentation, which is a dual supervised deep model with multi-dimensional attention and cascade multi-scale convolution mechanisms. In $\omega$-net, we first add an additional expansive path to bring additional learning loss, strengthen the supervision signal, and enhance the deep model's capability in restoring lost information. Then we add multi-dimensional attention (MDA) into this model, different from the conventional attention mechanism which use only xxxx attention, MDA applies both spatial position and channel attentions to xxxx (advantages). Finally, we introduce cascade multi-scale convolution (Cascade MSC) operations into the skip connections to alleviate the disparity between the concatenated features.
Using solely skip connections may not be sufficient to fully recover the lost information, and inevitably results in inaccurate segmentation for small objects. Therefore, the need of a more accurate deep model for medical image segmentation is compelling.

Consequently, in this work, we propose a novel dual supervised medical image segmentation model with multi-dimensional attention and cascade multi-scale convolution, which is called $\omega$-Net and shown in Fig.~\ref{fig_1}. So our deep model enhances the learning capability in the expansive path restoring location information. Secondly, we add multi-dimensional attention (MDA) into this model. Then, this deep model can capture more important object features with multi-dimension. Thirdly, we introduce cascade multi-scale convolution (Cascade MSC) operations into the skip connections.

\subsection{Dual Supervised Segmentation with An Additional Expansive Path}
\label{sec_EEP}

The first improvement of $\omega$-Net is to incorporate an additional expansive path into U-Net to achieve more accurate medical image segmentation by dual supervision. The up-sampled feature maps in each layer of original expansive path are not only concatenated with the feature maps from contracting path but also with the up-sampled feature maps from additional expansive path. Finally, with the help of additional expansive path, $\omega$-Net is learned by dual supervision, i.e., by both the segmentation loss from the original expansive path and the auxiliary segmentation loss from the additional expansive path.

This will make the segmentation learning in original expansive path take into account not only the feature information from contracting path, but also consider the intermediate segmentation information from additional expansive path. Therefore, we can treat the segmentation process in the additional expansive path as a coarse segmentation, whose coarse intermediate segmentation results are then sent to the original expansive path for further refinement, so as to obtain more accurate final segmentation results in original expansive path.

%Generally, medical images are first fed into the contracting path (i.e., the down sampling path in the right of Fig.~\ref{fig 1}) to generate contracting feature maps. In the contracting path, we encode the input image by five-layer operations of convolution and max-pooling. We build our convolution layers with asymmetric convolution blocks for better performance~\cite{ding2019acnet}, whose outputs are produced in every layer by a block consisted of two consecutive element-wise sum operations of the feature maps from the kernel filters with the sizes of $3 \times 3$, $1 \times 3$, and $3 \times 1$, and a max-pooling operation.

First, by denoting the input image as X, the output feature map at the $i^{th}$ (where $i>1$) layer of contracting path is written as
	\begin{equation}
		\begin{aligned}
			\textbf{F}_i=Pool\_Max(Conv\_2(\textbf{F}_{i-1})).
		\end{aligned}
	\end{equation}
\noindent where $Conv\_2()$ means two consecutive convolutional operations, and $Pool\_Max()$ denotes the max-pooling operation. Given $F_d$ (where $d$ is the number of layers in contracting path) as the most abstract feature map, the feature map can be formally written as
	\begin{equation}
		\begin{aligned}
			\textbf{Fu}^a_1=Trans\_Conv(\textbf{F}_d).
		\end{aligned}
	\end{equation}
\noindent Then with the skip-connection operation, the feature map generated by the $j^{th}$ ($j>1$) transposed-convolution-based up-sampling in additional expansive path can be formally written as
	\begin{equation}
		\begin{aligned}
			\textbf{Fu}^a_j=Trans\_Conv(Concate(\textbf{Fu}^a_{j-1},\textbf{F}_{d-j})),
		\end{aligned}
	\end{equation}
\noindent where $Concate()$ denotes the concatenation operation.

Similarly, the feature map generated by the first transposed-convolution-based up-sampling in original expansive path can be formally defined as
	\begin{equation}
		\begin{aligned}
			\textbf{Fu}^o_1=Trans\_Conv(\textbf{F}_d).
		\end{aligned}
	\end{equation}
\noindent And the feature map generated by the $j^{th}$ (where $j> 1$) transposed-convolution-based up-sampling in original expansive path can be formally defined as
	\begin{equation}
		\begin{aligned}
			\textbf{Fu}^o_j=Trans\_Conv(Concate(\textbf{Fu}^o_{j-1}, Concate(\textbf{Fu}^a_{j-1},\textbf{F}_{d-j}))).
		\end{aligned}
	\end{equation}

Finally, the auxiliary segmentation loss at the additional expansive path is formally defined as
	\begin{equation}
		\begin{aligned}
			L_a=BCE(Conv^{64 \times L}(\textbf{Fu}^a_{d-1}), Mask),
		\end{aligned}
	\end{equation}
\noindent where $L$ is the number of channels of the given annotations, $Conv^{64 \times L}()$ is a convolutional operation whose number of input channel is $64$ and number of output channel is $L$. Similarly, the segmentation loss at the original expansive path can be formally written
	\begin{equation}
		\begin{aligned}
			L_s=BCE(\textbf{Fu}^o_{d-1}, Mask).
		\end{aligned}
	\end{equation}
\noindent Consequently, the dual supervision loss of $\omega$-Net is formally
	\begin{equation}
		\begin{aligned}
			L_{dual}= \lambda \mathcal{L}_{a}+\mathcal{L}_{s},
		\end{aligned}
	\end{equation}
\noindent where $\lambda$ is a hyper-parameter which controls the relative importance of $L_a$ and $L_s$ in the dual supervision loss.

\subsection{Multi-Dimensional Self-Attention (MDSA)}
%\subsection{Spatial Position Attention (SPA)}
%\label{sec_SPA}
Although skip connections have been used in U-Net to remedy this problem by concatenating up-sampled feature maps in expansive path with the corresponding feature maps in contracting path to try to fill up the missing information.

Therefore, in order to resolve this problem, Attention U-Net~\cite{oktay2018attention} has been proposed to exploit attention gates in skip connection to suppress irrelevant regions in the concatenated feature maps from the contracting path while highlighting the salient features that are useful for specific segmentation tasks. Therefore, to avoid this problem, in $\omega$-Net, we propose to utilize a multi-dimensional self-attention (MDSA) mechanism to import weight information into the features on concatenated feature maps. As shown in Figure~\ref{fig_1}, multi-dimensional self-attention (MDSA) blocks are added and only added into skip connections between the additional expansive path and the original expansive path; this is to make the segmentation model capable of conducting self-attention operations on all feature maps that are sent to original expansive path while avoid redundant computations.

With the help of dense spatial position attention (DSPA), the features of small objects that are not salient on the feature map can now be enhanced using the salient features on the dense feature matrix that are highly similar to them, even if the salient features are extracted from regions that are far away from the small objects on the feature map. As shown in Fig.~\ref{fig_1}, the input feature map of DSPA, $\textbf{M} \in \mathbb{R}^{C\times H\times W}$, generated at the $j^{th}$ layer of the additional expansive path is the concatenation of feature maps at the $j^{th}$ layer of the additional expansive path, $\textbf{Fu}^a_{j}$, and those from the corresponding layer of the contracting path, $\textbf{F}_{d-j-1}$. Formally,
	\begin{equation}%\label{eq10}
		%\begin{split}
		\textbf{M}=Concate(\textbf{Fu}^a_{j},\textbf{F}_{d-j-1}),
		%\end{split}
	\end{equation}
\noindent where $d$ is the total number of layers of the contracting path.

We first reshape the input feature map \textbf{M} to $\mathbb{R}^{C\times N}$, where $N$ is the total number of pixels in each channel. \textbf{M} is then sent into a dilated convolution block to generate a dense feature matrix $\textbf{D} \in \mathbb{R}^{C\times K}$, where $K$ is a hyper-parameter representing the number of dense features in each channel of $\bf{D}$. Formally, 
	\begin{equation}%\label{eq10}
		%\begin{split}
		a_{j,i}=\dfrac{exp(\textbf{D}_i \cdot \textbf{M}_j)}{\sum_{i=1}^{N} exp(\textbf{D}_i \cdot \textbf{M}_j)},
		%\end{split}
	\end{equation}
\noindent where $a_{j,i}$ is an element of the dense spatial attention matrix $\bf{A}$ locating at the $j^{th}$ raw and the $i^{th}$ column, measuring the impact of the $i^th$ feature of the dense feature matrix $\bf{D}$ on the $j^{th}$ feature of the input feature map $\bf{M}$. Finally, we reshape the summation result to get the final output feature map $\textbf{O} \in \mathbb{R}^{C\times H\times W}$ of DSPA. Formally,
  \begin{equation}%\label{eq10}
		%\begin{split}
		\textbf{O}_{j}=\sum_{i=1}^{N} (a_{j,i}\textbf{D}_i) + \textbf{M}_j.
		%\end{split}
	\end{equation}

Similarly, a channel attention (CA) module is then used in MDSA to capture the channel dependencies between any two channel maps using a similar self-attention mechanism, where each channel map is updated by summarizing all weighted channel maps.

The detailed operations and formal definitions of channel attention (CA) are as follows. As shown in Fig.~\ref{fig_1}, given an input feature map of CA, $\textbf{M}' \in \mathbb{R}^{C\times H\times W}$, 
we first reshape \textbf{M}' to $\mathbb{R}^{C\times N}$, and then perform a matrix multiplication between the transpose of \textbf{M} and $\bf{M}$', and use a softmax operation to calculate the channel attention matrix $\textbf{A}' \in \mathbb{R}^{C\times C}$. Formally, 
	\begin{equation}%\label{eq10}
		%\begin{split}
		a'_{j,i}=\dfrac{exp(\textbf{M}'_i \cdot \textbf{M}'_j)}{\sum_{i=1}^{N} exp(\textbf{M}'_i \cdot \textbf{M}'_j)},
		%\end{split}
	\end{equation}
where $a_{j,i}$ is an element of the channel attention matrix $\textbf{A}'$ locating at the $j^{th}$ raw and the $i^{th}$ column, measuring the impact of the $i^th$ channel on the $j^{th}$ channel.

A matrix multiplication is conducted between $\bf{M}'$ and the transpose of $\bf{A}'$, whose result is then added to $\bf{M}'$ using a element-wise sum operation. Finally, we reshape the summation result to get the final output feature map $\textbf{O}' \in \mathbb{R}^{C\times H\times W}$ of CA. Formally,
  \begin{equation}%\label{eq10}
		%\begin{split}
		\textbf{O}'_{j}=\sum_{i=1}^{N} (a'_{j,i}\textbf{M}'_i) + \textbf{M}'_j.
		%\end{split}
	\end{equation}

\subsection{Cascade Multi-Scale Convolution (Cascade MSC)}

The performances of U-Net based segmentation models may be weakened if we concatenate these feature maps directly using skip connections. Some advanced U-Net variants propose to resolve this problem by adding some convolution operations on the skip connection, among which U-Net++ is the state-of-the-art solution. The feature maps generated in the deeper layer is more abstract, so they are usually more likely to lose some important information of small objects; Therefore, using this strategy to fuse multi-scale feature information may not be appropriate for the small object segmentation task.

Therefore, in this work, we propose a new multi-scale feature fusion solution, cascade multi-scale convolution (cascade MSC). These multi-scale feature maps are then fused to bridge the semantic gaps. Since all the multi-scale feature maps generated by cascade MSC at a given skip connection are based on the same source feature maps, and does not use any feature maps from deeper layer, the feature information of small objects can be retained in the generated multi-scale feature maps to a greater extent.

Formally,

	\begin{equation}
		\begin{aligned}
			\textbf{X}^1&= Conv\_5\times 5(\textbf{X})),\\
			\textbf{X}^2&= Conv\_3\times 3(Sum(\textbf{X},\textbf{X}^1)),\\
			\textbf{X}^3&= Conv\_1\times 1(Sum(\textbf{X},\textbf{X}^2)),\\
			\textbf{Y}&= Conv\_1\times 1(Concate(\textbf{X},\textbf{X}^1,\textbf{X}^2,\textbf{X}^3)).\\
		\end{aligned}
	\end{equation}

\section{Experiments}
\label{Experiments}
\subsection{Experimental Setup}
\label{sec_ES}
%We introduce our experimental setup, including three used real-world datasets and pre-processing, implementation, and baselines.

\subsubsection{Datasets and Pre-processing}
We evaluate the proposed $\omega$-Net model on the following three computerized tomography (CT) image datasets, where the size of images is $512 \times 512$.

\textbf{Kidney Data}~\cite{heller2019kits19}, \textbf{Pancreas Data}~\cite{simpson2019large} and \textbf{Liver Data}~\cite{simpson2019large} are divided into three subsets for training (70\%), validation (10\%), and testing (20\%). The mask of targets (organs and its tumor) is generated based on manual annotation before training the networks. The mask of the manual annotation is applied as our label rather than the original annotation image. Then, the two channels' values from the annotation calculated as ground truth compared with the output prediction.

\subsubsection{Implementation}
Our proposed method is conducted with the backbone of U-Net. The filter-groups number of layers in encoder branch is set $\left[64, 128, 256, 512,1024\right]$ respectively and the kernel size is $3$, while the number of filters in decoder branch is set as $\left[1024,512, 256, 128, 64\right]$. The size of the deepest latent representation vector is $1024 \times 32 \times 32$. In the position attention module, we set the value of $K$ as $10$. The number of filters of the output layer is set as $2$, and the kernel size is $1$. The Adam optimizer with a weight decay of $0.00015$ is applied in initialization for decreasing overfitting. $\lambda_{out}$ is set as $10$ and $\lambda_m$ is set as $1$. To minimize the variance of each observation as small as possible to get more stable convergence, we adopt gradient accumulation on the platform Pytorch within the limited memory. All the networks are evaluated by using the public platform Pytorch.

\subsubsection{Baselines}
Extensive experiments are conducted to compare our proposed method with state-of-the-art baselines. to prove the effectiveness of our proposed MDA$^*$-U-Net$^*$, we conduct FCN~\cite{long2015fully}, U-Net~\cite{ronneberger2015u}, Attention U-Net~\cite{oktay2018attention} (i.e., utilizing attention gates in the decoder branch of U-Net to automatically suppress the response of the irrelevant background and enhance the sensitivity for foreground information.), and U-Net$++$~\cite{zhou2018unet++} (i.e., reducing the semantic gap between the feature maps of the encoder and decoder by nested dense convolution operations.) as the baselines. 

\subsubsection{Evaluation metrics}
The spatial consistency between the automated segmentation and the ground truth with the manual reference segmentation was quantified by the Dice Similarity Coefficient (DSC), Positive Predictive Value (PPV), and Sensitivity. They are calculated according to equations as follows:

\begin{small}
	\begin{equation}
	DSC = \frac{2TP}{T+P}, \qquad PPV = \frac{TP}{TP+FP}, \qquad Sensitivity=\frac{TP}{TP+FN},
	\end{equation}
\end{small}

\noindent where Positives (P) are samples that belong to organ and tumor, Negatives (N) are samples that belong to the background, True (T) demonstrate correct pixel-wise classification for segmentation, and False (F) means mistake on predicting segmentation.

\subsection{Main Results}
\label{sec_MP}
We evaluate the performances of our proposed method on three datasets, which are compared with the state-of-the-art baselines, including FCN, U-Net, Attention U-Net and U-Net$++$. As is shown in Fig.~\ref{fig 2}, $\omega$-Net designed to equip with additional auxiliary loss and is implemented by the architecture composed of contracting path, expansive path, cascade multi-scale convolution modules and multi-dimensional attention module consisted of spatial position attention block and channel attention block. In our proposed $\omega$-Net, we get the resulting feature maps in each layer by concatenating two types of feature maps: \emph{up-sampled feature maps} and \emph{attentive feature maps}. Up-sampled feature maps are obtained by applying the transposed convolution operation on feature maps in the previous layers; while the \emph{muti-dimensional attention} module generates attentive feature maps. Especially, multi-dimensional attention (MDA) module takes the corresponding feature maps in the original expansive path as inputs, and its operations are as follows: It first applies a two-dimensional (2D) multiplication operation on the tiled feature maps for position attention to strengthen the deep model's learning on areas with interested small object features. Then, making the output of the position attention module as input, other two-dimensional (2D) multiplication operation is used on the tiled feature maps for channel attention to enhance the deep model's learning capability where the feature representation of small objects are delicate. 

%%%%%%Visualization: Comparison on ...
\begin{figure}
	%\vspace{-0mm}
	\includegraphics[width=\linewidth]{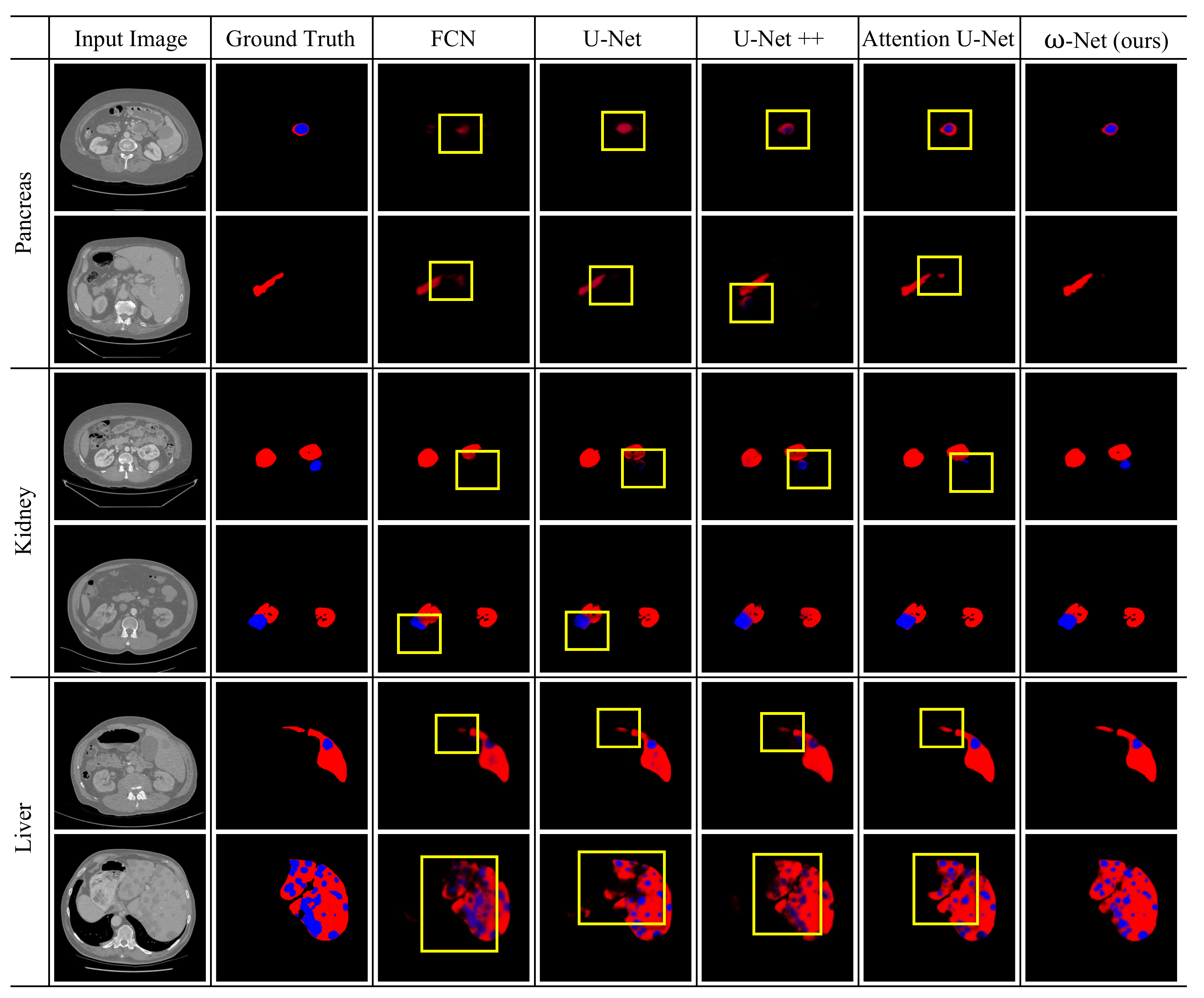}
	\centering
	\caption{\small The evaluation comparison of our proposed \emph{MDA$^*$-U-Net$^*$} and the other four baselines on three CT datasets.}
	%\vspace{-0mm}
	\label{fig 3}
\end{figure}

%%%%%%%%%%%%%%%%%%%%%%% %[width=0.9\linewidth,cols=4,pos=h]
\begin{table*}
	\centering
	\caption{Performances of the proposed MDA$^*$-U-Net$^*$ and the state-of-the-art baselines.}
	\label{Table 1}
	\begin{tabular}{c|ccc|ccc|ccc}
		\toprule
		\multirow{2}{*}{Achitecture} & \multicolumn{3}{c|}{Kidney} & \multicolumn{3}{c|}{Pancreas} & \multicolumn{3}{c}{Liver}\\
		\cmidrule{2-10}
		&DSC & PPV & Sensi & DSC & PPV & Sensi & DSC & PPV & Sensi \\
		%\midrule
		\cmidrule{1-10}
		FCN & 0.8811 & 0.9113 & 0.8551 & 0.7514 & 0.8023 & 0.7316 & 0.9197 & 0.9277 & 0.9074 \\
		U-Net & 0.8969 & 0.9362 & 0.8657 & 0.8205 & 0.8365 & 0.8123 & 0.9257 & 0.9365 & 0.9256 \\
		U-Net$++$ & 0.9096 & 0.9401 & 0.8771 & 0.8221 & 0.8591 & 0.8176 & 0.9273 & 0.9381 & 0.9269 \\
		Attention U-Net & 0.9204 & 0.9421 & 0.9024 & 0.8345 & 0.8526 & 0.8205 & 0.9319 & 0.9402 & 0.9298 \\
		MDA$^*$-U-Net$^*$ & \bf 0.9380 & \bf 0.9540 & \bf 0.9325 & \bf 0.8540 & \bf 0.8803 & \bf 0.8363 & \bf 0.9492 & \bf 0.9516 & \bf 0.9392 \\
		\cmidrule{1-10}
		Improvement & \bf 0.0176 & \bf 0.0119 & \bf 0.0301 & \bf 0.0195 & \bf 0.0212 & \bf 0.0158 & \bf 0.0173 & \bf 0.0114 & \bf 0.0094 \\
		\bottomrule 
	\end{tabular}
\end{table*}
%%%%%%%%%%%%%%%%%%%%%%%%%%%%%%%%%%%%%%%%%%%%%%%%

\section{Discussion}
\label{Discussion}
In this section, we first summarize our proposed \emph{MDA$^*$-U-Net$^*$} method. After that, we also point out the main differences between our proposed multi-dimension attention module and other attention methods, cascade multi-scale convolution module and the previous studies on multi-scale learning approaches. Finally, we briefly state the advantages of our proposed method.

\subsection{Summary on Our Proposed Method}
Our proposed \emph{MDA$^*$-U-Net$^*$} method is built by introducing an additional \emph{multi-dimensional attention based expansive path} into U-Net and further integrating the model with \emph{cascade multi-scale convolution (Cascade MSC)} operations (i.e., used on the skip connection and MDA module). Firstly, the resulting feature maps in each contracting layer is fed as input into the cascade multi-scale convolution block, then concatenating output \emph{dense multi-scale feature maps} and \emph{up-sampled feature maps}. Finally, the expansive path predict segmentation by: concatenating \emph{up-sampled feature maps} and \emph{attentive feature maps}, and then generating outputs through transposed convolution operations.

\subsection{Comparison with Previous Work}
Compared with the existing deep-learning-based segmentation methods in the literature~\cite{ni2019raunet},~\cite{hu2018squeeze}, our proposed method has the different and improved architecture (i.e., including multi-dimensional attention module and cascade multi-scale convolution module.) Our proposed multi-dimensional attention method integrate deep semantic spatial position attention information and feature channel attention maps into a unified segmentation framework. In this way,  to enhance the small object learning capability, these two correlated sub-attention modules are exploited to adaptively coordinate the model's attention optimization on deep semantic analysis level. In addition, rather than using solely the mono-scale feature representations, our proposed method extracted and fused dense multi-scale feature maps~\cite{zhao2017pyramid} to construct delicate representations addressing the small object segmentation. To this end, we introduce the cascade multi-scale convolution module in the segmentation framework.

In Table.~\ref{Table 1}, we present the results of several state-of-the-art baselines reported in the literature for medical images segmentation using the methods end-to-end, including four related encode-decode methods. According to those methods evaluation on the above three datasets for DSC, PPV, and Sensitivity, we can have at least two observations. First, our method consistently leads to competitive performance. Moreover, it's worth noting that our proposed method obtains more improvement on the kidney and pancreas datasets than the liver CT images. It may prove that our proposed method can especially solve the small objects segmentation problem better than general segmentation task. The liver region account for a larger rate of area than both kidney and pancreas images, and it locate at the simpler abdominal surrounding (i.e., on the contrary, the kidney and pancreas surround complex similar characteristic abdominal organs tissue).

\section{Conclusion}
\label{Conclusion}
Accurate segmentation is an essential step of radiotherapy, and its automation and precision improvement is worthy of investigation. In this study, we propose $\omega$-Net to automatically predict medical image segmentation. On the three public datasets (i.e., kidney CT images with 300 patients, pancreas CT images with 283 cases, and liver CT images with 183 cases), the effectiveness of our proposed method on small objects segmentation has been extensively evaluated. Compared with several state-of-the-art baselines, our proposed method has demonstrated better segmentation performance, especially on the pancreas and its tumor segmentation task.

\appendix

\section*{Acknowledgments}
\noindent This work was supported in part by the National Natural Science Foundation of China under the grant 61906063, in part by the Natural Science Foundation of Tianjin City, China, under the grant 19JCQNJC00400, in part by the ``100 Talents Plan'' of Hebei Province, China, under the grant E2019050017, and in part by the Yuanguang Scholar Fund of Hebei University of Technology, China.

\bibliographystyle{cas-model2-names}

\bibliography{cas-refs}
\end{document}